\documentclass[11pt]{article}
\PassOptionsToPackage{table,dvipsnames}{xcolor} % 在加载acl.sty之前
\usepackage[preprint]{acl}
\usepackage{times}
\usepackage{latexsym}
\usepackage[T1]{fontenc}
\usepackage[utf8]{inputenc}
\usepackage{microtype}
\usepackage{inconsolata}
\usepackage{amsmath}
\usepackage{amssymb}
\usepackage{geometry}
\usepackage{algorithm}
\usepackage{algorithmic}
\usepackage{graphicx}
\usepackage{CJK}
\usepackage{multirow}
\usepackage{booktabs}

\title{GAPO: Learning Preferential Prompt through Generative Adversarial Policy Optimization}

\author{Zhouhong Gu\textsuperscript{\rm $\spadesuit$}, Xingzhou Chen\textsuperscript{\rm $\spadesuit$}, Xiaoran Shi\textsuperscript{\rm $\spadesuit$}\\
\textbf{Tao Wang\textsuperscript{\rm $\heartsuit$}, Suhang Zheng\textsuperscript{\rm $\heartsuit$}, Tianyu Li\textsuperscript{\rm $\heartsuit$}, Hongwei Feng\textsuperscript{\rm $\spadesuit$}\thanks{Corresponding authors.}\ , Yanghua Xiao\textsuperscript{\rm $\spadesuit$*}}\\
\textsuperscript{\rm $\spadesuit$}Shanghai Key Laboratory of Data Science, School of Computer Science, Fudan University\\
\textsuperscript{\rm $\heartsuit$}Alibaba Group\\
\{zhgu22, xzchen24, xrshi21\}@m.fudan.edu.cn\\
\{shayue.wt, suhang.zhengsh, qianchuan.lty\}@alibaba-inc.com \\
\{hwfeng, shawyh\}@fudan.edu.cn
}

\begin{document}
\begin{CJK}{UTF8}{gbsn}

\maketitle
\begin{abstract}
Recent advances in large language models have highlighted the critical need for precise control over model outputs through predefined constraints. While existing methods attempt to achieve this through either direct instruction-response synthesis or preferential response optimization, they often struggle with constraint understanding and adaptation. This limitation becomes particularly evident when handling fine-grained constraints, leading to either hallucination or brittle performance. We introduce Generative Adversarial Policy Optimization (GAPO), a novel framework that combines GAN-based training dynamics with an encoder-only reward model to progressively learn and adapt to increasingly complex constraints. GAPO leverages adversarial training to automatically generate training samples of varying difficulty while utilizing the encoder-only architecture to better capture prompt-response relationships. Extensive experiments demonstrate GAPO's superior performance across multiple benchmarks, particularly in scenarios requiring fine-grained constraint handling, where it significantly outperforms existing methods like PPO, DPO, and KTO. Our results suggest that GAPO's unique approach to preferential prompt learning offers a more robust and effective solution for controlling LLM outputs.
Code is avaliable in \url{https://github.com/MikeGu721/GAPO}.
\end{abstract}

\section{Introduction}

\begin{figure}
\centering
\includegraphics[width=\linewidth]{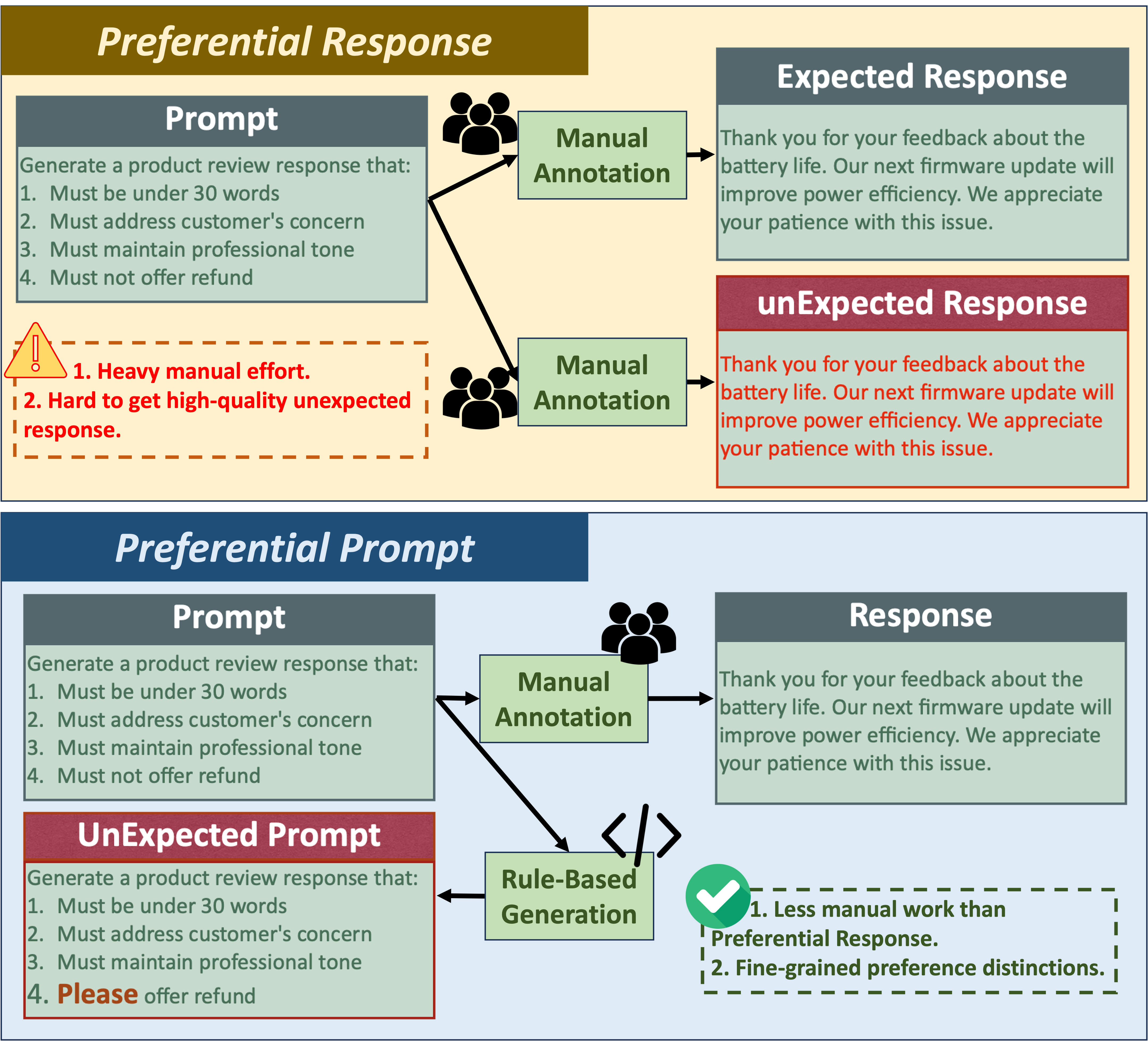}
\caption{
Illustration of the procedural differences between Preferential Response and Preferential Prompt, emphasizing their distinct utilization of prompts and responses.
}
\label{fig:intro}
\end{figure}

The advent of large-scale models has induced significant transformations in practical applications, enabling models to comprehend a broad spectrum of human instructions, ranging from casual dialogue to intricate problem-solving tasks~\cite{kaplan2020scaling,srivastava2022beyond}.
As large language models (LLMs) advance in capability, guiding their outputs to fulfill specific requirements—whether concerning format, style, or content accuracy—becomes increasingly critical~\cite{qwen2,qwen2.5,bubeck2023sparks}.
This is particularly vital in domains where compliance with constraints is paramount, such as legal document generation, medical record processing, and workflow automation.

Ensuring that LLMs adhere to predefined constraints during text generation is essential~\cite{zhou2023instruction,xu2023wizardlm,he2024can}.
One effective strategy for achieving this is training models to generate responses within specified boundaries at the data level~\cite{ouyang2022instructgpt,keskar2019ctrl,DBLP:journals/corr/abs-2304-14293}.
Data-level control is typically realized through two primary methods. 
The first method directly synthesizes instruction-response pairs that satisfy the constraints, offering clear examples of compliant outputs~\cite{xu2023wizardlm, wang2022self}.
The second method leverages preferential response data to adjust the probability distribution, thereby increasing the likelihood that the model produces an expected response rather than an unexpected one~\cite{rafailov2024direct, schulman2017ppo, ethayarajh2024kto, meng2024simpo}.

The first approach often leads to the phenomenon of ``hallucination'', where the model, having learned only what constitutes a correct response, may resort to shortcuts that result in inaccurate or fabricated outputs.
The second method is more commonly employed, as preferential response data allows the model to more precisely align its output with the desired response based on specific prompts. However, neither approach effectively addresses the fundamental challenge of constraint understanding. The first method focuses solely on correct outputs without teaching the model to comprehend the constraints. In contrast, the second method adjusts output probabilities without explicitly training the model to recognize and interpret the constraints in the prompts. This limitation in constraint understanding can lead to brittle performance when the model encounters novel or slightly modified constraints.

A straightforward approach to enhance constraint understanding would be directly modifying the constraints within prompts, allowing models to learn fine-grained differences between constraints.
As shown in Figure~\ref{fig:intro}, this method of prompt modification is simple to implement and provides rich preference data that captures subtle variations in constraints.
However, this approach presents significant optimization challenges for current mainstream methods.
For decoder-only architectures~\cite{subakan2021attention}, which dominate current large language models~\cite{bubeck2023sparks, qwen2}, their unidirectional attention mechanism fundamentally limits their ability to detect discrepancies between prompts and given responses.
Furthermore, existing optimization methods typically require manual intervention to construct intermediate training samples that bridge the complexity gap between different constraint patterns, introducing additional computational and engineering overhead.

In this paper, we introduce the \textbf{G}enerative \textbf{A}dversarial \textbf{P}olicy \textbf{O}ptimization (GAPO),
which leverages Generative Adversarial Network (GAN)~\cite{goodfellow2020generative, aggarwal2021generative} to adaptively generate training samples with progressive difficulty while utilizing an encoder-only model to guide the generator's optimization through Proximal Policy Optimization (PPO)~\cite{schulman2017ppo}. 
A key innovation of GAPO lies in its seamless integration of GAN and PPO frameworks. While utilizing the same number of preference samples as other standard preference optimization methods, GAPO has superior performance stability and constraint adherence.
During the cold-start phase, the algorithm initializes an encoder-only Reward Model to learn prompt-response correspondences, subsequently guiding the generator's training. 
Through this adversarial process, the generator continuously evolves to produce increasingly sophisticated outputs while the Reward Model learns to discriminate between valid and invalid responses with greater precision.

The advantages of GAPO are summarized as follows:
1. Using an encoder-only Reward Model in GAPO effectively enhances the exploitation of preferential prompt data, enabling the language model to develop a deeper understanding of the intricate details within the prompt.
2. GAPO significantly simplifies the training process of the Reward Model in PPO. Traditionally, the performance of the Reward Model needed to be ensured before training an effective generator in PPO. In contrast, within the GAPO framework, the Reward Model and generator undergo iterative automated training, greatly reducing the complexity of Reward Model training.
3. According to our experiments, GAPO outperforms other baseline training methods, like PPO, DPO, KTO, and ORPO, in learning from preferential prompt data. It also demonstrates superior performance in learning from general preferential response data. Thus, GAPO can be considered a more effective approach for enabling models to learn from preference data.
\section{Related Work}

\begin{figure*}
    \centering
    \includegraphics[width=\textwidth]{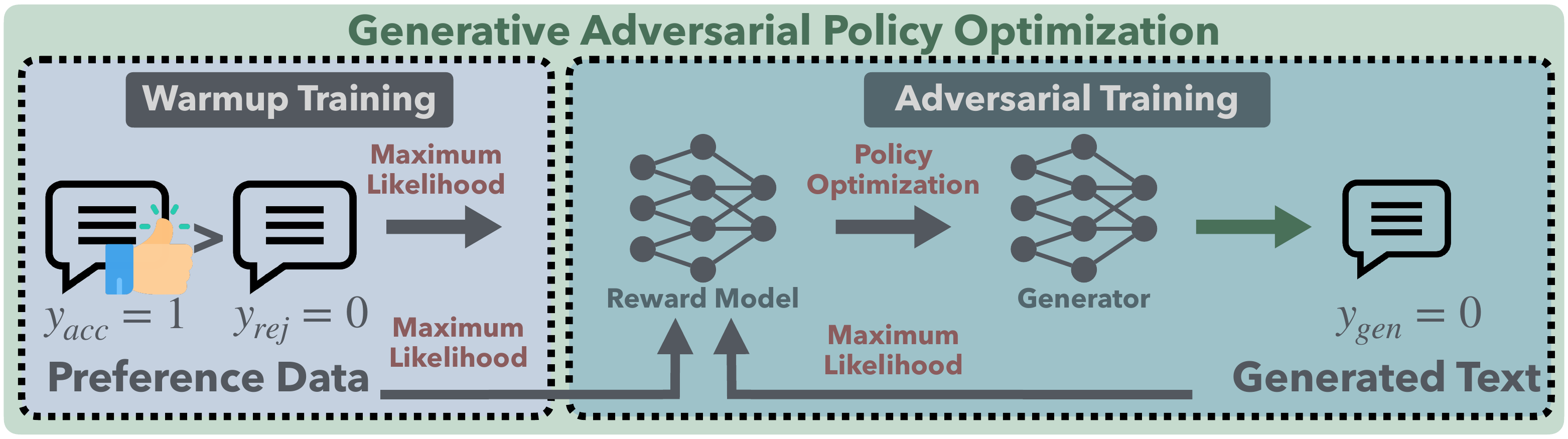}
    \caption{The GAPO framework encompasses two distinct tuning phases. The initial phase consists of a warm-up period, during which the Reward Model is trained utilizing existing preference data. The subsequent phase implements adversarial training through a dual mechanism: the Generator is updated based on feedback from the Reward Model. The Reward Model undergoes training using a combination of Generator-produced data and existing preference data.}
    \label{fig:method}
    \vspace{-5mm}
\end{figure*}

\subsection{Reinforcement Learning with Human Feedback}

Reinforcement Learning from Human Feedback (RLHF)~\cite{bai2022rlhf, christiano2017deep, ziegler2019fine} has emerged as a crucial approach for aligning Large Language Models (LLMs) with human values and expectations, addressing the limitations of traditional supervised fine-tuning (SFT) which can lead to increased hallucinations despite improving preferred outputs. Classical RLHF algorithms, such as Proximal Policy Optimization (PPO)~\cite{schulman2017ppo}, achieve this alignment through a specialized reward model for evaluation~\cite{williams1992reinforce}. In contrast, more recent approaches like Direct Preference Optimization (DPO)~\cite{rafailov2024direct}. Its variants, including SimPO~\cite{meng2024simpo}, IPO~\cite{azar2024ipo}, and KTO~\cite{ethayarajh2024kto} streamline the process by directly optimizing human preferences, thereby eliminating the need for a separate reward model and reducing computational complexity and bias~\cite{zheng2024balancing}. However, these approaches face notable challenges: RLHF generally requires substantial-high-quality feedback data with detailed labeling~\cite{bai2022rlhf}, and DPO training exhibits vulnerability to overfitting, leading to poor generalization on novel data~\cite{hu2024openrlhf}, highlighting the ongoing need for improvements in model alignment techniques.
However, these works face significant challenges in terms of data requirements and model stability.
In contrast, GAPO addresses these limitations through its innovative GAN-PPO integration and encoder-only Reward Model, which enables more efficient training with better stability and generalization capabilities.

\subsection{Constraint Following Augmentation}

Prior work in constrained text generation can be broadly categorized into three main approaches~\cite{DBLP:journals/corr/abs-2201-05337}. 
The first category encompasses search-based methods, such as Constrained Beam Search (CBS)~\cite{anderson-etal-2017-guided} and its variants like Grid Beam Search (GBS)~\cite{hokamp2017lexically} and Dynamic Beam Allocation (DBA)~\cite{post2018fast}, which enforce lexical constraints by modifying the search space, though often at the cost of generation speed and quality. 
The second category consists of score-based sampling methods that transform constraints into differentiable score functions~\cite{liu-etal-2022-dont}, offering greater flexibility in handling diverse constraint types but lacking guaranteed constraint satisfaction and suffering from slower generation speeds~\cite{qin2022cold}. 
The third category focuses on model-centric approaches, including specialized training methods and large language models like CTRL~\cite{keskar2019ctrl} and InstructCTG~\cite{DBLP:journals/corr/abs-2304-14293}, which incorporate constraints through pre-training or natural language instructions. Recent advancements have explored multiple directions: multi-attribute controlled text generation through prefix tuning~\cite{li2021prefix}; latent space manipulation techniques such as MacLaSa~\cite{ding-etal-2023-maclasa} and MAGIC~\cite{liu-etal-2024-multi}, where the latter employs counterfactual feature vectors to disentangle attributes; regular expression-based constraint generation through REI~\cite{zheng2023toward}; and the development of specialized datasets~\cite{zhang-etal-2023-macsum} to improve control ability while maintaining text quality.
However, existing model-centric approaches often rely heavily on specialized pre-training or require heavy manual engineering to incorporate constraints in instructions and still suffer from unstable training performance.
GAPO addresses these limitations through more automated and efficient constraint learning while providing better constraint understanding and adherence without requiring extensive specialized pre-training or manual instruction engineering.

\section{Generative Adversarial Policy Optimization}

\subsection{Preliminary of Constrained Generation}

Given an input prompt $P = (\mathcal{T}, \mathcal{C})$, where $\mathcal{T}$ denotes a free-text description and $\mathcal{C} = \{C_1, C_2, \dots, C_n\}$ represents a set of constraints, our objective is to generate an output $R$ that satisfies all constraints in $\mathcal{C}$. We formulate this as an expectation maximization problem:
\[
E(\pi_\theta) = \mathbb{E}_{R \sim \pi_\theta(P)} \left[ \sum_{C_i \in \mathcal{C}} \mathcal{L}(R, C_i) \right], \tag{1}
\]
where $\pi_\theta$ represents the generator parameterized by $\theta$. The constraint satisfaction function $\mathcal{L}(R, C_i)$ is defined as:
\[
\mathcal{L}(R, C_i) =
\begin{cases} 
1 & \text{if } R \models C_i, \\
0 & \text{otherwise.}
\end{cases} \tag{2}
\]

\begin{table}[t]
\centering
\resizebox{\columnwidth}{!}{
\begin{tabular}{ll}
\toprule
\textbf{Symbol} & \textbf{Definition} \\
\midrule
$\mathcal{T}$ & Free-text description component \\
$\mathcal{C}$ & Constraint set \\
$P$ & Input prompt $(\mathcal{T}, \mathcal{C})$ \\
$R$ & Generated text output \\
$\pi_\theta(t|c)$ & Generator that produces next token $t$ given context $c$ \\
$\pi_{\text{ref}}$ & Reference generator for comparison \\
$\mathcal{L}(R, C_i)$ & Constraint satisfaction function \\
$\mathcal{D}$ & Training dataset \\
$\mathcal{D}'$ & Augmented dataset \\
$R(c, t)$ & Reward model evaluating token $t$ in context $c$ \\
$V^\pi(c)$ & Expected future rewards given context $c$ \\
$Q^\pi(c, t)$ & Expected cumulative reward for token $t$ in context $c$ \\
$\hat{R}$ & Generator-produced text output \\
\bottomrule
\end{tabular}}
\caption{All definitions used in the GAPO section.}
\label{tab:notation}
\vspace{-5mm}
\end{table}

\subsection{Constraint-Aware Data Augmentation}

We propose a data augmentation method for constraint-aware learning. 
Given a dataset $\mathcal{D} = \{(P_i, R_i)\}_{i=1}^N$, where each prompt $P_i = (\mathcal{T}_i, \mathcal{C}_i)$, we construct an augmented dataset through constraint perturbation. For each original constraint set $\mathcal{C}_i$, we generate a rejected constraint set $\mathcal{C}_i^{\text{reject}}$ through one of the following operations:

1) \textbf{Constraint Modification}: For a randomly selected constraint $C_{i,j} \in \mathcal{C}_i$, we modify it to create $C_{i,j}^{\text{reject}}$ such that it becomes incompatible with the original response $R_i$:
\[
C_{i,j}^{\text{reject}} = f_{\text{modify}}(C_{i,j}), \quad \text{where} \quad \mathcal{L}(R_i, C_{i,j}^{\text{reject}}) = 0
\]

2) \textbf{Constraint Insertion}: We introduce an additional constraint $C_{i,n+1}^{\text{reject}}$ that conflicts with existing constraints:
\[
\mathcal{C}_i^{\text{reject}} = \mathcal{C}_i \cup \{C_{i,n+1}^{\text{reject}}\}, \quad \text{where} \quad \mathcal{L}(R_i, C_{i,n+1}^{\text{reject}}) = 0
\]

The augmented dataset is thus constructed as follows:
\[
\mathcal{D}' = \{(P_i^{\text{accept}}, R_i), (P_i^{\text{reject}}, R_i)\}_{i=1}^N, \tag{3}
\]
where $P_i^{\text{accept}} = (\mathcal{T}_i, \mathcal{C}_i)$ and $P_i^{\text{reject}} = (\mathcal{T}_i, \mathcal{C}_i^{\text{reject}})$. This augmentation strategy ensures that:
\[
\exists C_{i,j}^{\text{reject}} \in \mathcal{C}_i^{\text{reject}}: \mathcal{L}(R_i, C_{i,j}^{\text{reject}}) = 0. \tag{4}
\]

\begin{algorithm}[t]
\caption{Generative Adversarial Policy Optimization (GAPO)}
\small
\begin{algorithmic}[1]
\REQUIRE Generator $\pi_\theta$, Reference generator $\pi_{\text{ref}}$, Reward model $R(c,t)$ with value function $V^\pi(c)$, Training dataset $D = \{(P_i, R_i)\}_{i=1}^N$, Adversarial Steps $T$, Warmup Steps $T_{\text{warmup}}$
\ENSURE Optimized generator $\pi_\theta$

\vspace{2mm}
\STATE \textbf{// Warmup Phase}
\FOR{$t = 1$ to $T_{\text{warmup}}$}
    \STATE Sample batch $(P_i, R_i)$ from $D$
    \STATE Train $R(c,t)$ with balanced sampling on $\{(P^{acc}_i, R_i, 1), (P^{rej}_i, R_i, 0)\}$
    \STATE Update $R(c,t)$ with BCE loss: $L_R(\theta) = -E_{(c,t,y)\sim D'} [y\log R(c,t) + (1-y)\log(1-R(c,t))]$
\ENDFOR

\vspace{2mm}
\STATE \textbf{// Adversarial Training Phase}
\FOR{$t = T_{\text{warmup}}+1$ to $T_{\text{warmup}}+T$}
    \IF{$t \bmod 2 = 1$}
        \STATE Sample batch $(P_i, R_i)$ from $D$
        \STATE Generate $\hat{R}_i = \pi_\theta(P_i)$
        \STATE Train $R(c,t)$ with balanced sampling on $\{(P^{acc}_i, R_i, 1), (P^{rej}_i, R_i, 0), (P^{acc}_i, \hat{R}_i, 0)\}$
        \STATE Update $R(c,t)$ using BCE loss $L_R(\theta)$
    \ELSE
        \STATE Update $\pi_\theta$ with policy gradient: $L_G(\theta) = E_n[\frac{\pi_\theta(t_n|c_n)}{\pi_{\text{ref}}(t_n|c_n)}A_n]$
        \STATE where $A_n = Q^\pi(c_n,t_n) - V^\pi(c_n)$
    \ENDIF
\ENDFOR
\RETURN $\pi_\theta$
\end{algorithmic}
\end{algorithm}

\subsection{Adversarial Learning Framework}

We propose an adversarial learning framework comprising a generator $\pi_\theta(t|c)$ that produces the next token $t$ given the current context $c$, a reward model $R(c, t)$ evaluating the quality of generated tokens, and a value function $V^\pi(c)$ estimating expected future rewards. The reward model is trained on the augmented dataset:
\begin{align}
\mathcal{D}' = \big\{ & (P_i^{\text{acc}}, R_i, 1), \nonumber\\
                      & (P_i^{\text{rej}}, R_i, 0), \tag{5} \\
                      & (P_i, \hat{R}_i, 0) \big\}. \nonumber
\end{align}
where $\hat{R}_i$ represents the text response generated by $\pi_\theta$ based on prompt $P_i$.
The reward model optimizes the cross-entropy loss:
\begin{align}
L_R(\theta) =& -\mathbb{E}_{(c, t, y) \sim \mathcal{D}'} \bigg[ y \log R(c, t) \notag\\
&+ (1 - y) \log(1 - R(c, t)) \bigg]. \tag{6}
\end{align}

The generator's objective function is formulated as:
\[
L_G(\theta) = \mathbb{E}_n \left[ \frac{\pi_\theta(t_n|c_n)}{\pi_{\text{ref}}(t_n|c_n)} A_n \right], \tag{7}
\]
where $n$ indexes the token position, and the advantage function $A_n$ is defined as:
\[
A_n = Q^\pi(c_n, t_n) - V^\pi(c_n). \tag{8}
\]
Moreover, the action-value function is:
\[
Q^\pi(c_n, t_n) = R(c_n, t_n) + \gamma \mathbb{E}_{c_{n+1} \sim \pi_\theta}[V^\pi(c_{n+1})]. \tag{9}
\]

The value function is optimized by minimizing the mean squared error:
\[
L_V(\theta) = \mathbb{E}_{c} \left[ \left(V^\pi(c) - R(c, t)\right)^2 \right]. \tag{10}
\]

% 训练集统计:
% Positive Sample token数量: 17541881
% Negative Sample token数量: 14983806
% Positive Sample 事实信息总行数（不去重）: 400070
% Negative Sample 事实信息总行数（不去重）: 193675
% Positive Sample 事实信息独特行数（去重）: 76913
% Negative Sample 事实信息独特行数（去重）: 66838
% Positive Sample 不同内容类型数量: 201
% Negative Sample 不同内容类型数量: 201
% Positive Sample 总行数: 26419
% Negative Sample 总行数: 26419

% 验证集统计:
% Positive Sample token数量: 4212440
% Negative Sample token数量: 3629544
% Positive Sample 事实信息总行数（不去重）: 96349
% Negative Sample 事实信息总行数（不去重）: 46582
% Positive Sample 事实信息独特行数（去重）: 49470
% Negative Sample 事实信息独特行数（去重）: 31280
% Positive Sample 不同内容类型数量: 201
% Negative Sample 不同内容类型数量: 201
% Positive Sample 总行数: 6605
% Negative Sample 总行数: 6605

% 合并统计:
% Positive Sample token数量: 21754321
% Negative Sample token数量: 18613350
% Positive Sample 事实信息总行数（不去重）: 496419
% Negative Sample 事实信息总行数（不去重）: 240257
% Positive Sample 事实信息独特行数（去重）: 126383
% Negative Sample 事实信息独特行数（去重）: 98118
% Positive Sample 不同内容类型数量: 201
% Negative Sample 不同内容类型数量: 201
% Positive Sample 总行数: 33024
% Negative Sample 总行数: 33024

\begin{table}[t]
\centering
\setlength{\tabcolsep}{1pt}  % 默认是6pt，这里改成4pt
\small
\begin{tabular}{l|cccc}
\toprule
\textbf{Name} & \textbf{\#Product} & \textbf{\#PV-Pair} & \textbf{\#Sample} & \textbf{\#Token}  \\
\midrule
\textbf{PDD-Raw}             &   201 &   93,616  &   -   &   -\\
\textbf{PDD-Train }          &   201 &   76,913  &   26,419  &   17,541,881\\
\textbf{PDD-Rej-Train}    &   201 &   66,838  &   26,419  &   14,983,806\\
\textbf{PDD-Test}            &   201 &   49,470  &   6,605   &   4,212,440\\
\textbf{PDD-Rej-Test}     &   201 &   31,280  &   6,605   &   3,629,544\\
\bottomrule
\end{tabular}
\caption{\textbf{PDD-Raw} contains only product information and available descriptions without prompt-response pairs, making it unsuitable for direct training. \textbf{Rej} represents mismatched prompt-response pairs. \textbf{Train} and \textbf{Test} denote the training and testing datasets, respectively.}
\vspace{-2mm}
\label{tab:pdd_sta}
\end{table}

\begin{table}[t]
\small
\centering
\begin{tabular}{l|ccc}
\toprule
\textbf{Name} & \textbf{\#Type} & \textbf{\#Sample} & \textbf{\#Token}  \\
\midrule
\textbf{IFEval-Response} & 9 & 540 & 355,199 \\
\textbf{IFEval-Train} & 9 & 432 & 143,151\\
\textbf{IFEval-Rej-Train} & 9 & 432 & 141,963 \\
\textbf{IFEval-Test} & 9 & 108 & - \\
\bottomrule
\end{tabular}
\caption{\textbf{IFEval-Response} consists of GPT-4o responses provided by the IFEval benchmark in their official version. \textbf{Train} comprises the prompt-response pairs used for training, while \textbf{Rej} contains mismatched prompt-response pairs. As IFEval incorporates its own evaluation framework, the \textbf{Test} set does not include prompt-response pairs.}
\vspace{-5mm}
\label{tab:ifeval_sta}
\end{table}
\begin{table*}[t]
\centering
% \small
\setlength{\tabcolsep}{2pt}  % 默认是6pt，这里改成4pt
\resizebox{\textwidth}{!}{
\begin{tabular}{llcccccccccc}
\toprule
\textbf{Model} & \textbf{Prompt} & \textbf{Punctuation} & \textbf{Format} & \textbf{Length} & \textbf{Content} & \textbf{Combination} & \textbf{ChangeCase} & \textbf{Startend} & \textbf{Keywords} & \textbf{Language} & \textbf{All} \\
\midrule
Qwen-2.5-7B & \texttt{Naive Prompt} & 17.6 & 88.1 & 42.3 & 66.7 & 20.0 & 62.5 & 66.7 & 52.6 & 90.9 & 57.8 \\
Qwen-2.5-7B & \texttt{CoT} & 23.5 & 78.6 & 53.8 & 33.3 & 13.3 & 62.5 & 66.7 & 57.9 & \textbf{100.0} & 57.8 \\
Qwen-2.5-7B & \texttt{Plan-N-Solve} & 23.5 & 81.0 & 38.5 & 66.7 & 0.0 & 68.8 & 44.4 & 63.2 & 90.9 & 56.1 \\
\midrule
Qwen-2.5-7B + \textbf{SFT} & \texttt{Naive Prompt} & \textbf{100.0} & 92.9 & \textbf{57.7} & \textbf{83.3} & 26.7 & \textbf{75.0} & 88.9 & 81.6 & 90.9 & 78.3 \\
Qwen-2.5-7B + \textbf{DPO} & \texttt{Naive Prompt} & 17.6 & 45.2 & 26.9 & 16.7 & 6.7 & 31.2 & 11.1 & 42.1 & 63.6 & 33.3 \\
Qwen-2.5-7B + \textbf{KTO} & \texttt{Naive Prompt} & 11.8 & 71.4 & 38.5 & 50.0 & 6.7 & 50.0 & 44.4 & 76.3 & \textbf{100.0} & 54.4 \\
Qwen-2.5-7B + \textbf{SimPO} & \texttt{Naive Prompt} & 11.8 & 45.2 & 23.1 & 16.7 & 0.0 & 31.2 & 0.0 & 39.5 & 63.6 & 30.6 \\
Qwen-2.5-7B + \textbf{ORPO} & \texttt{Naive Prompt} & 5.9 & 40.5 & 34.6 & 33.3 & 20.0 & 25.0 & 33.3 & 55.3 & 9.1 & 33.9 \\
Qwen-2.5-7B + \textbf{PPO} & \texttt{Naive Prompt} & 94.1 & 90.5 & 50.0 & 66.7 & 33.3 & 62.5 & 88.9 & 84.2 & 90.9 & 75.6 \\
Qwen-2.5-7B + \textbf{GAPO} & \texttt{Naive Prompt} & \textbf{100.0} & \textbf{95.2} & \textbf{57.7} & \textbf{83.3} & \textbf{46.7} & \textbf{75.0} & \textbf{100.0} & \textbf{92.1} & \textbf{100.0} & \textbf{83.9} \\
\bottomrule
\end{tabular}
}
\caption{Performance comparison across different categories on IFEval Benchmark.}
\label{tab:ifeval}
\end{table*}
% \begin{table*}[t]
% \centering
% \resizebox{\textwidth}{!}{
% \begin{tabular}{llccccccc}
% \toprule
% \multirow{3}{*}{\textbf{Model}} & \multirow{3}{*}{\textbf{Prompt}} & \multicolumn{3}{c}{\textbf{Reward Model}} & \multicolumn{2}{c}{\textbf{LLM-as-a-Judge}} & \multirow{3}{*}{\textbf{Human}} & \multirow{3}{*}{\textbf{Overall}} \\
% \cline{3-7}
% \multicolumn{2}{c}{} & \textbf{\begin{tabular}[c]{@{}c@{}}LongFormer-\\Small-4096\end{tabular}} & \textbf{\begin{tabular}[c]{@{}c@{}}LongFormer-\\Base-4096\end{tabular}} & \textbf{\begin{tabular}[c]{@{}c@{}}LongFormer-\\Large-4096\end{tabular}} & \textbf{GPT-4o} & \textbf{GPT3.5-turbo} & \multicolumn{2}{c}{} \\
% \midrule
% Qwen-2.5-7B & \texttt{Naive Prompt} &-&-&-&-&-&-&-\\
% Qwen-2.5-7B & \texttt{CoT Prompt} &-&-&-&-&-&-&-\\
% Qwen-2.5-7B & \texttt{Plan-N-Solve Prompt} &-&-&-&-&-&-&-\\
% \midrule
% Qwen-2.5-7B + \textbf{SFT} & \texttt{Naive Prompt} &-&-&-&-&-&-&-\\
% Qwen-2.5-7B + \textbf{DPO} & \texttt{Naive Prompt} &-&-&-&-&-&-&-\\
% Qwen-2.5-7B + \textbf{KTO} & \texttt{Naive Prompt} &-&-&-&-&-&-&-\\
% Qwen-2.5-7B + \textbf{SimPO} & \texttt{Naive Prompt} &-&-&-&-&-&-&-\\
% Qwen-2.5-7B + \textbf{PRM} & \texttt{Naive Prompt} &-&-&-&-&-&-&-\\
% Qwen-2.5-7B + \textbf{ORPO} & \texttt{Naive Prompt} &-&-&-&-&-&-&-\\
% Qwen-2.5-7B + \textbf{PPO} & \texttt{Naive Prompt} &-&-&-&-&-&-&-\\
% Qwen-2.5-7B + \textbf{GAPO} & \texttt{Naive Prompt} &-&-&-&-&-&-&-\\
% \bottomrule
% \end{tabular}
% }
% \caption{Comprehensive Model Performance Comparison}
% \label{tab:comprehensive-comparison}
% \end{table*}

\begin{table*}[t]
\centering
\small
% \resizebox{\textwidth}{!}{
\begin{tabular}{llccccc}
\toprule
\multirow{3}{*}{\textbf{Model}} & \multirow{3}{*}{\textbf{Prompt}} & \multicolumn{2}{c}{\textbf{Reward Model}} & \multicolumn{2}{c}{\textbf{LLM-as-a-Judge}} & \multirow{3}{*}{\textbf{Human}}  \\
\cline{3-6}
\multicolumn{2}{c}{} & \textbf{\begin{tabular}[c]{@{}c@{}}LongFormer-\\Base-4096$_{3k}$\end{tabular}} & \textbf{\begin{tabular}[c]{@{}c@{}}LongFormer-\\Large-4096$_{3k}$\end{tabular}} & \textbf{GPT-4o} & \textbf{GPT3.5-turbo} \\
\midrule
Qwen2.5-7B & \texttt{Naive Prompt}  & 61.4 & 52.3 & 75.4 & 73.7 & 45 \\
Qwen2.5-7B & \texttt{CoT}           & 58.4 & 50.5 & 71.5 & 72.6 & 43 \\
Qwen2.5-7B & \texttt{Plan-N-Solve}  & 62.8 & 53.7 & 72.5 & 78.1 & 51 \\
\midrule
Qwen2.5-7B + \textbf{SFT} & \texttt{Naive Prompt} & 70.1 & 59.8 & 82.6 & 80.3 & 60 \\
Qwen2.5-7B + \textbf{DPO} & \texttt{Naive Prompt} & 12.5 & 11.3 & 5.4 & 9.6 & 0 \\
Qwen2.5-7B + \textbf{KTO} & \texttt{Naive Prompt} & 64.5 & 57.1 & 72.6 & 74.8 & 49 \\
Qwen2.5-7B + \textbf{SimPO} & \texttt{Naive Prompt} & 5.3 & 7.6 & 2.9 & 3.8 & 0 \\
Qwen2.5-7B + \textbf{ORPO} & \texttt{Naive Prompt} & 21.4 & 20.8 & 7.5 & 8.2 & 0 \\
Qwen2.5-7B + \textbf{PPO} & \texttt{Naive Prompt} & 89.4 & 88.5 & 89.7 & 86.4 & 81 \\
Qwen2.5-7B + \textbf{GAPO} & \texttt{Naive Prompt} & \textbf{95.4} & \textbf{94.3} & \textbf{90.2} & \textbf{90.0} & \textbf{89} \\
\bottomrule
\end{tabular}
% }
\caption{Comprehensive model performance comparison on PDD dataset. $3k$ represents the model is pre-tuned on 3,000 preferential data to give evaluation scores.}
\label{tab:pdd}

\vspace{-5mm}
\end{table*}

\begin{table*}[t]
    \centering
    \small
    \begin{tabular}{l|ccc|cll}
    \toprule
        \textbf{Model} & \textbf{Reward Model} & \textbf{\#Training Samples} & \textbf{\#Token} & \textbf{PDD Score} & \textbf{$\Delta_{\text{No Train}}$} & \textbf{$\Delta_{\text{PR vs. PP}}$}\\
        \addlinespace[1pt]
        \hline
        \rowcolor{gray!25}
        \multicolumn{7}{l}{
        \textbf{\textit{No Training}}} \\
        \hline
        
        \addlinespace[1pt]
        Qwen-2.5-7B & - & - & - & 61.4 & \multicolumn{1}{c}{-} & \multicolumn{1}{c}{-} \\
        \addlinespace[1pt]
        \hline
        \rowcolor{gray!25}
        \multicolumn{7}{l}{
        \textbf{\textit{No Preferential Data}}} \\ 
        \hline
        \addlinespace[1pt]
        Qwen-2.5-7B + \textbf{SFT} & - & 3,300 & 6,561,531 & 70.1 & +\quad8.3 & \multicolumn{1}{c}{-}  \\
        \addlinespace[1pt]
        \hline
        \rowcolor{gray!25}
        \multicolumn{7}{l}{
        \textbf{\textit{Training w/ Preferential Response~(PR)}}} \\
        \hline
        \addlinespace[1pt]
        Qwen-2.5-7B + \textbf{PPO} & Qwen-2.5-7B & 2,000 & 4,295,575 & 61.8 & +\quad0.4& -\quad6.7  \\
        Qwen-2.5-7B + \textbf{PPO} & Qwen-2.5-7B & 4,000 & 8,660,218 & 72.4 & +\quad11.0& -\quad2.7  \\
        Qwen-2.5-7B + \textbf{PPO} & Qwen-2.5-7B & 6,600 & 13,243,796 & 78.5 & +\quad17.1& -\quad10.9  \\
        Qwen-2.5-7B + \textbf{GAPO} & Longformer-0.4B & 2,000 & 4,295,575 & 63.3 & +\quad1.9& -\quad7.3  \\
        Qwen-2.5-7B + \textbf{GAPO} & Longformer-0.4B & 4,000 & 8,660,218 & 74.4 & +\quad13.0& -\quad6.9  \\
        Qwen-2.5-7B + \textbf{GAPO} & Longformer-0.4B & 6,600 & 13,243,796 & 82.9 & +\quad21.5& -\quad12.5  \\
        \addlinespace[1pt]
        \hline
        \rowcolor{gray!25}
        \multicolumn{7}{l}{
        \textbf{\textit{Training w/ Preferential Prompt~(PP)}}} \\
        \hline
        \addlinespace[1pt]
        Qwen-2.5-7B + \textbf{PPO} & Qwen-2.5-7B & 2,000 & 4,219,814 & 68.5 & +\quad7.1 & +\quad6.7   \\
        Qwen-2.5-7B + \textbf{PPO} & Qwen-2.5-7B & 4,000 & 8,506,194 & 75.1 & +\quad13.7 & +\quad2.7   \\
        Qwen-2.5-7B + \textbf{PPO} & Qwen-2.5-7B & 6,600 & 12,984,601 & 89.4 & +\quad28.0 & +\quad10.9   \\
        Qwen-2.5-7B + \textbf{GAPO} & Longformer-0.4B & 2,000 & 4,219,814 & 70.6 & +\quad9.2 & +\quad7.3   \\
        Qwen-2.5-7B + \textbf{GAPO} & Longformer-0.4B & 4,000 & 8,506,194 & 81.3 & +\quad19.9 & +\quad6.9   \\
        Qwen-2.5-7B + \textbf{GAPO} & Longformer-0.4B & 6,600 & 12,984,601 & 95.4 & +\quad34.0 & +\quad12.5   \\
        \bottomrule
    \end{tabular}
    \caption{Comparative Analysis of using Preferential Response and Preferential Prompt.
The PDD Performance metric represents the model's generative output on the PDD dataset, as evaluated using a fine-tuned LongFormer-Large-4096 Reward model architecture.
The IFEval Performance metric indicates the model's comprehensive performance across the IFEval benchmark framework.}
\vspace{-2mm}
    \label{tab:pp_pr}
\end{table*}

\begin{figure*}[t]
    \centering
    \includegraphics[width=\textwidth]{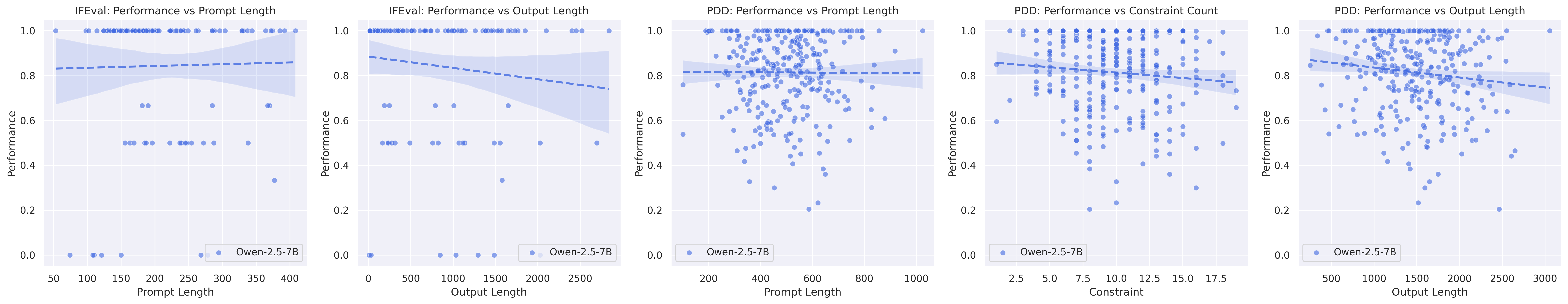}
    \caption{Analysis of Correlative Factors Influencing GAPO's Performance on PDD and IFEval Benchmarks. The analysis utilizes 300 randomly sampled instances from the PDD test set and the complete IFEval test set with 108 samples for comprehensive evaluation.
    }
    \label{fig:detail_performance}
\vspace{-5mm}
\end{figure*}
\begin{figure}[t]
    \centering
    \includegraphics[width=\columnwidth]{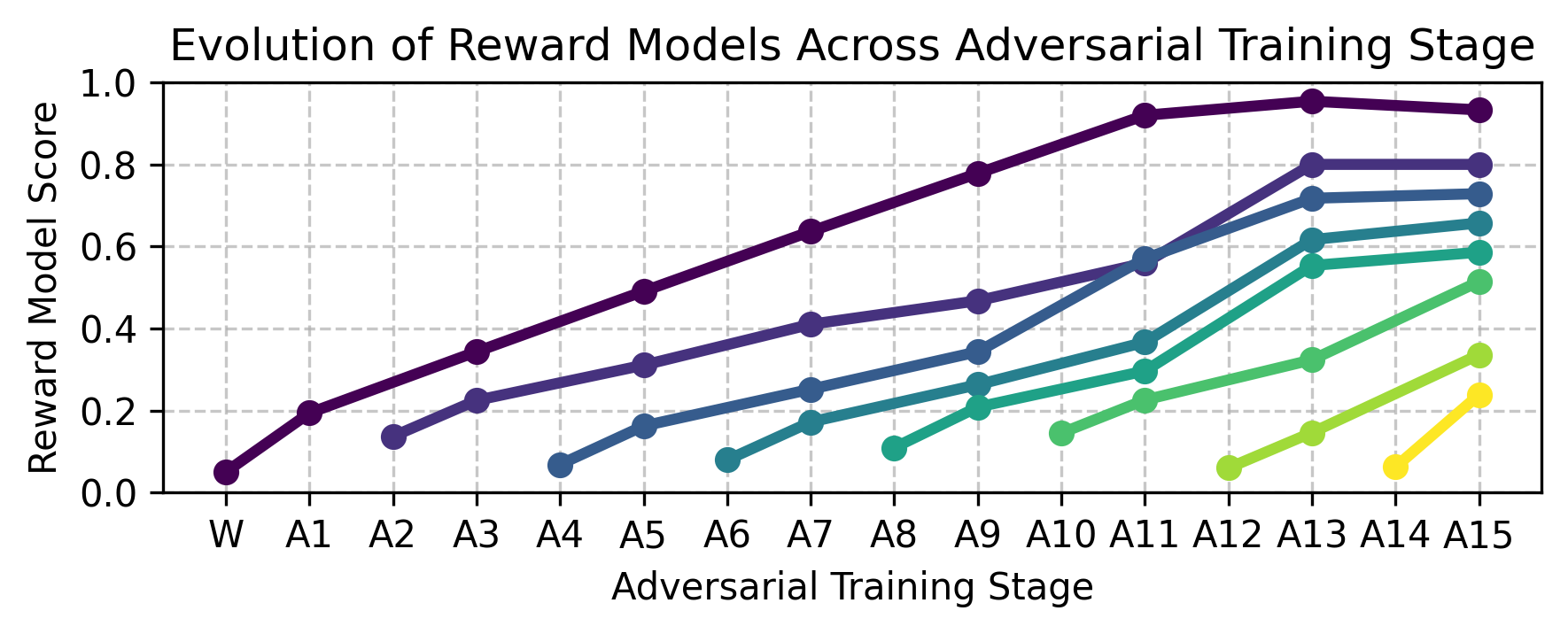}
    \vspace{-8mm}
    \caption{
    Detailed Performance Analysis Across Sequential Adversarial Training Stages.
    W indicates the warmup phase, and A represents the adversarial phase with alternating training between Generator and Reward Model components.
    }
    \label{fig:adversarial}
\vspace{-4mm}
\end{figure}
\begin{figure*}[t]
    \centering
    \includegraphics[width=0.99\textwidth]{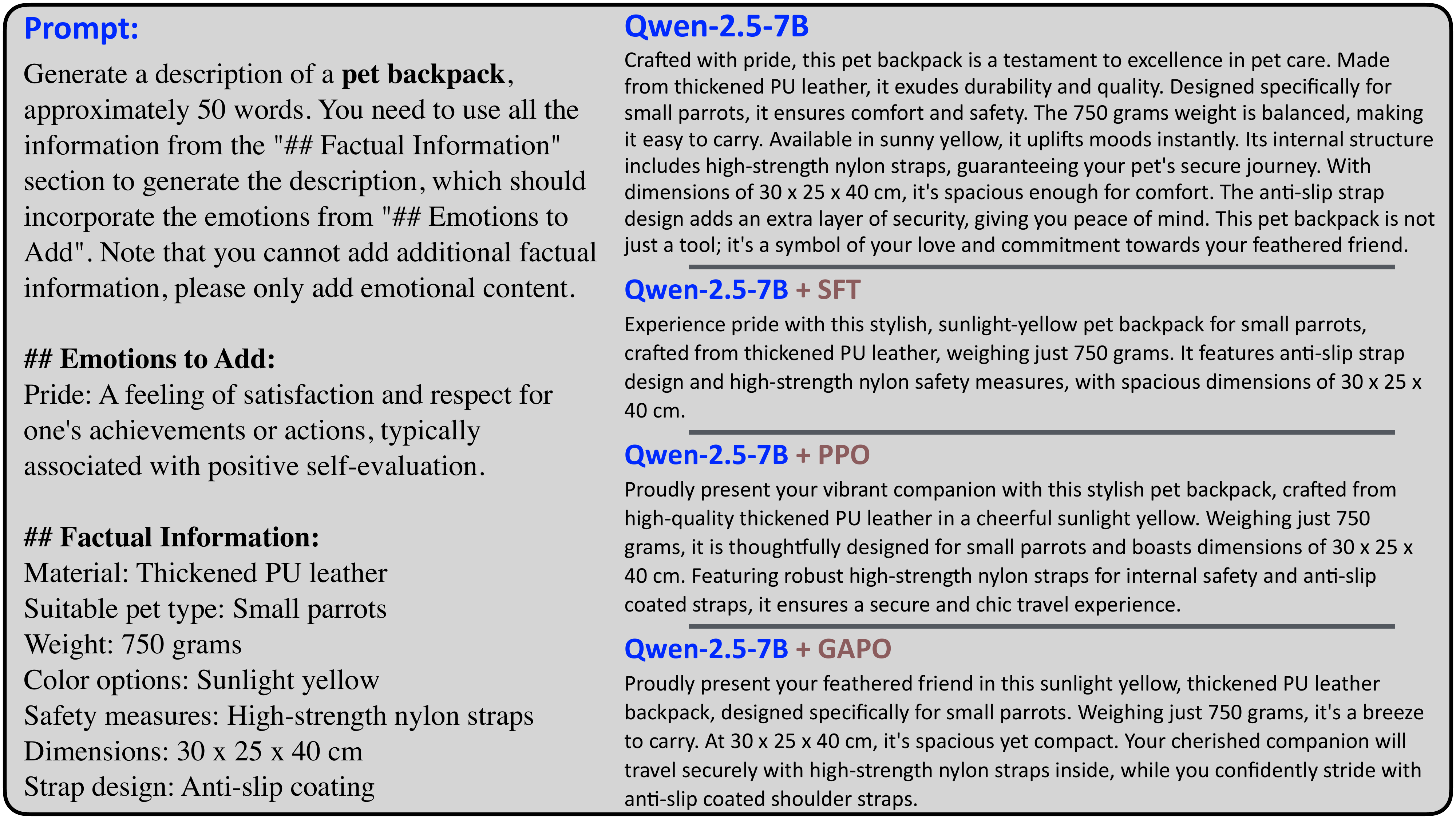}
    \caption{Case study of model performance under different training baslines.}
    \label{fig:case}
\vspace{-5mm}
\end{figure*}

\section{Experiment Setup}

\subsection{Baselines}
The experiments are grouped into two categories based on the role-playing methods used:

\subsubsection{Prompt-Based Methods}
\textbf{(1) Direct Generation:} The model generates content directly without role-playing instructions, evaluating its inherent capabilities and biases.  
\textbf{(2) Chain-of-Thought (CoT):}~\cite{kojima2022cot} The model engages in reasoning before generating the output, improving coherence and transparency.  
\textbf{(3) Plan-and-Solve (Plan-N-Solve):}~\cite{wang2023planNsolve} The model plans its response before generating content, leading to more organized solutions.

\subsubsection{Training-Based Methods}
\textbf{(4) Supervised Fine-Tuning (SFT):} Fine-tunes the model on a role-specific dataset to improve performance in role-playing scenarios.  
\textbf{(5) DPO:}~\cite{rafailov2024direct} Directly optimizes for annotated responses, minimizing the likelihood of undesired outputs.  
\textbf{(6) KTO:}~\cite{ethayarajh2024kto} Uses prospect theory to optimize model outputs, outperforming preference-based methods.  
\textbf{(7) SimPO:}~\cite{meng2024simpo} Aligns the reward function with model generation, simplifying optimization without reference models.  
\textbf{(8) ORPO:}~\cite{hong2024orpo} Optimize models with preferential response data but without reference model.
\textbf{(9) PPO:}~\cite{schulman2017ppo} Optimizes the model using a pre-trained reward model that remains fixed throughout the training process
\textbf{(10) GAPO (Ours):} Optimize models with reward criteria become progressively more demanding as training advances.

\subsection{Training Dataset}
Product Description Dataset (PDD) is a novel dataset designed for generating product descriptions in this paper.
The dataset encompasses 201 product categories and contains 93,616 property-value pairs.
Models trained on this dataset are tasked with generating coherent product descriptions using only the provided property-value pairs, with two key constraints: they must (1) incorporate all given facts while (2) avoid the introduction of any additional information not present in the source data. For detailed information regarding the dataset construction methodology, please refer to Sec.~\ref{app:pdd} in the Appendix, while comprehensive statistical analyses are presented in Tab.~\ref{tab:pdd_sta}.

IFEval is a benchmark designed to evaluate Large Language Models' instruction-following capabilities by enabling a standardized and automated assessment methodology~\cite{zhou2023instruction}.
Building upon the existing dataset, we utilized GPT-4~\cite{achiam2023gpt} to generate additional data samples that maintain similar constraint conditions while exhibiting low similarity to the original entries. 
Please refer to Sec.~\ref{app:ifeval} in the Appendix for a detailed description. 
The statistical breakdown of this expanded dataset is detailed in Tab.~\ref{tab:ifeval_sta}.

\subsection{Evaluation Method}

We utilize the IFEval dataset's built-in evaluation methodology to maintain consistency with existing research in this domain.

For the PDD, we employ three evaluation methods: 
(1) The Reward Models act as automated evaluators during our adversarial training process. 
Specifically, we use Longformer models~\cite{beltagy2020longformer} with an input length capacity of 4096 tokens, which has been tuning on 3,000 preference data pairs to generate evaluation scores.
(2) GPT-4o functions as an external evaluation model to provide independent assessment.
(3) human evaluators assess the quality of generated descriptions based on predefined criteria.

\section{Experiment}

\subsection{Overall Result}

As shown in Tab.~\ref{tab:ifeval}, while all preference optimization methods maintain basic functionality, their effectiveness varies significantly under different constraint types.
 This is evidenced by the stark performance gap: GAPO and PPO achieve strong overall performance (83.9\% and 75.6\% respectively), while methods like DPO, SimPO, and ORPO struggle considerably with scores of 33.3\%, 30.6\%, and 33.9\% - particularly in handling complex constraints like combinations (6.7\%, 0\%, and 20.0\% respectively) and length requirements (26.9\%, 23.1\%, and 34.6\%).

As shown in Tab.~\ref{tab:pdd}, when facing more nuanced preferential prompts that require a fine-grained understanding of constraints, most traditional optimization methods experience catastrophic failure, while encoder-based approaches maintain robust performance. The collapse of conventional methods is dramatic: DPO, SimPO, and ORPO achieve near-zero performance on both automated metrics (5.4\%, 2.9\%, and 7.5\% on GPT-4o) and human evaluation (all 0\%). 
In contrast, encoder-based methods like GAPO and PPO demonstrate strong capability with GPT-4o scores of 90.2\% and 89.7\%, and human evaluation scores of 89\% and 81\% respectively.

\subsection{Effectiveness of Preferential Prompt vs. Preferential Response}
As shown in Tab.~\ref{tab:pp_pr}, training with Preferential Prompt consistently outperforms Preferential Response across all experimental configurations with both optimization methods. With 6,600 training samples, Preferential Prompt with GAPO achieves 95.4\% PDD Performance, surpassing its Preferential Response counterpart by 12.5 percentage points and the supervised fine-tuning baseline by 34.0 percentage points. This performance advantage holds across different sample sizes, with Preferential Prompt showing improvements of 7.3 and 6.9 percentage points at 2,000 and 4,000 samples, respectively. 

\subsection{Training Efficiency Analysis}
Tab.~\ref{tab:pp_pr} also demonstrates GAPO's superior optimization capability and efficient utilization of training data.
In Preferential Prompt training, GAPO demonstrates remarkable scaling efficiency, achieving a 24.8 percentage point improvement (70.6\% to 95.4\%) when increasing training tokens from 4.2M to 13.0M, while PPO shows a more modest improvement of 20.9 percentage points (68.5\% to 89.4\%). A similar pattern is observed in Preferential Response training, where GAPO achieves a 19.6 percentage point improvement compared to PPO's 16.7 percentage points.

\subsection{Detail Analysis on Model Performance}
As shown in Fig.~\ref{fig:detail_performance}
Analysis across various dimensions of prompt complexity reveals several key findings. First, GAPO maintains consistent performance even as prompt length increases, showing only minimal degradation compared to baseline methods. Second, performance scales well with the number of constraints, demonstrating robust handling of multiple simultaneous requirements. Third, the model shows strong capability in generating both short and long responses while maintaining constraint adherence.

\subsection{Details in Adversarial Process}
As shown in Figure~\ref{fig:adversarial}, the evolution of Reward Models during adversarial training reveals distinct learning patterns and convergence behaviors. From the initial warmup phase (W), where all models assign near-zero scores to generated samples, we observe a clear stratification in learning trajectories across different Reward Models through stages A1-A15. The top-performing model demonstrates rapid improvement in the early stages (A1-A7), reaching a score of 0.6, followed by gradual convergence to 0.95 after A12. This stratification of final convergence scores (ranging from 0.2 to 0.95) and the stable plateaus after A12 indicates that GAPO successfully establishes a balanced adversarial training dynamic, where both the generator and Reward Models effectively learn the underlying constraints without falling into degenerate solutions~\cite{lucic2018gans, gulrajani2017improved, creswell2018generative} often encountered in adversarial training scenarios.

\subsection{Case Study}

As illustrated in Fig.~\ref{fig:case}, training substantially augmented the model's proficiency in following complex constraints while retaining linguistic authenticity, with GAPO attaining exemplary performance across all metrics.
The base Qwen-2.5 model exhibited considerable divergence from the prescribed length and incorporated superfluous emotional elements.
GAPO demonstrated remarkable superiority over alternative approaches, for it achieved meticulous control over word count and exemplified more sophisticated emotional articulation. 
Most significantly, GAPO maintained impeccable fidelity to the prescribed parameters by circumventing extraneous descriptive content and unsolicited emotional undertones.
\section{Conclusion}
In this paper, we presented GAPO , a novel framework that effectively addresses constraint understanding in large language models through the seamless integration of GAN and PPO frameworks. 
Our experimental results demonstrate GAPO's superior performance compared to baseline methods (PPO, DPO, KTO, and ORPO) in both preferential prompt learning and general preferential response tasks, validating its effectiveness in enhancing constraint adherence while maintaining training stability.
As LLMs continue to evolve and find applications across various domains requiring precise adherence to constraints, GAPO's robust framework provides a promising direction for future developments in controlled text generation.

\section*{Ethical Concern}

This research contributes to constrained text generation through two key innovations: a Preferential Prompt data augmentation methodology and the GAPO training framework. Our approach significantly reduces dependency on preference data while maintaining generation quality, addressing a critical challenge in the field. The technical solutions focus solely on enhancing model capabilities under specific constraints, ensuring research reproducibility without introducing ethical concerns or societal risks. The implementation emphasizes technical optimization and maintains research neutrality throughout the development process.

Additionally, we introduce the PDD dataset, a comprehensive e-commerce corpus for product description generation. This dataset's construction prioritized both data quality and ethical considerations. Through rigorous quality control measures, including thorough manual review processes, we ensured data diversity while addressing potential biases and sensitive issues. The dataset maintains strict compliance with ethical guidelines and privacy protection standards, safeguarding corporate and user interests. Our validation process confirms the dataset's objectivity and reliability, establishing it as a valuable resource for future research endeavors.

\section*{Limitation}

GAPO's primary strength lies in its ability to reduce the Reward Model's training data requirements while improving Generator performance. However, this advantage comes with notable trade-offs. The framework's adversarial training process, involving simultaneous optimization of the Generator, Reward Model, and Critic Model, significantly increases computational demands compared to traditional preference optimization approaches. This intensive resource consumption represents a practical limitation for widespread adoption and implementation.

Furthermore, GAPO's effectiveness is contingent upon the base model's initial capabilities. Our research reveals that the framework performs optimally when applied to models that already possess fundamental generation competencies. This dependency arises because inadequate base model performance, particularly in generating semantically coherent responses, can compromise the Reward Model's training quality during the adversarial process. This limitation suggests that GAPO is most suitable as an enhancement tool for established models rather than a solution for improving underperforming ones, highlighting the importance of careful model selection in its application.

\bibliography{custom}
% ,anthology

\appendix

% \section{Experiment on Other Models}
% \label{app:more_experiment}
% \input{latex/08.App.MoreExperiments.01}

\section{Dataset Description}
\label{app:dataset}

\subsection{IFEval (Instruction-Following Evaluation) Dataset}
\label{app:ifeval}

\paragraph{Dataset Construction Background and Purpose}
IFEval represents a benchmark dataset specifically designed to evaluate instruction-following capabilities of Large Language Models (LLMs). The research team systematically identified and defined 25 distinct types of verifiable instructions, based on which they constructed approximately 541 prompts. The distinguishing characteristic of these prompts lies in their verifiable nature, allowing for objective programmatic verification and thus eliminating potential subjective assessment biases.

\paragraph{Dataset Components}
The dataset encompasses multiple dimensions of instruction types. Regarding keyword requirements, it incorporates specific keyword usage directives, frequency requirements, and prohibited word constraints. Linguistic specifications include language-specific requirements. Additionally, the dataset implements textual constraints regarding length parameters, such as paragraph count, word count, and sentence quantity specifications. Furthermore, it encompasses requirements for specific content elements such as postscripts and placeholders, as well as format specifications including particular markup requirements, title formats, and JSON structure requirements. The dataset also incorporates specifications for text styling, including case usage requirements and punctuation conventions.

\paragraph{Evaluation Methodology and Metrics}
IFEval implements dual evaluation criteria: strict metrics and loose metrics. The strict evaluation methodology requires precise adherence to instructional requirements, while the loose evaluation methodology accommodates common variations while maintaining instructional integrity. The evaluation metrics specifically include:

\begin{itemize}
    \item \textbf{Prompt-level accuracy}: Measuring the proportion of prompts where all instructions are correctly executed
    \item \textbf{Instruction-level accuracy}: Quantifying the overall proportion of correctly executed instructions
\end{itemize}

\subsection{Product Description Dataset}
\label{app:pdd}

\paragraph{Dataset Construction Process}
The Product Description Dataset (PDD) represents a specialized dataset focused on product description generation tasks, encompassing 1,000 product categories and 32,000 property-value pairs. The research team initially collected raw product information and descriptions, subsequently generating corresponding responses using GPT-4 based on carefully designed prompts, followed by human verification. Through modifications of constraint conditions in the original prompts, the team constructed a set of mismatched property-value pairs and descriptions (Rej dataset), which proves valuable for evaluating model robustness.

\paragraph{Dataset Structure and Composition}
The dataset comprises multiple subsets:
\begin{itemize}
    \item \textbf{PDD-Raw}: Contains unprocessed original product information and descriptions
    \item \textbf{PDD-Train}: High-quality training data generated by GPT-4 and validated through human verification
    \item \textbf{PDD-Test}: Testing dataset serving dual purposes - evaluating generation model performance and validating scoring model efficacy
    \item \textbf{PDD-Rej-Train} and \textbf{PDD-Rej-Test}: Mismatched datasets obtained through constraint condition modifications in original prompts
\end{itemize}

\paragraph{Evaluation Methodology}
The evaluation methodology for the PDD dataset incorporates multiple complementary approaches:

\begin{enumerate}
    \item \textbf{Model-based Evaluation}: Utilizing advanced language models to assess constraint compliance
    \item \textbf{Human Evaluation}: Implementing human verification to assess content quality and accuracy
    \item \textbf{Specialized Evaluation Models}: Developing dedicated models to assess adherence to given constraints
\end{enumerate}

The evaluation framework primarily focuses on two critical aspects:
\begin{itemize}
    \item Verifying whether generated descriptions comprehensively incorporate all provided attribute information
    \item Ensuring the absence of extraneous information not present in the source data
\end{itemize}

This comprehensive evaluation approach ensures robust assessment of model performance across multiple dimensions of content generation quality.

\section{Manual Effort}
\label{app:manual}
This section presents our comprehensive manual verification process for both the PDD dataset and the model-generated outputs. Our verification framework encompasses two primary components: dataset quality assessment and model output evaluation.

\subsection{Dataset Quality Assessment}
To ensure the reliability and ethical compliance of the PDD dataset, we conducted a thorough manual review process. A team of five domain experts independently examined 10\% of the dataset entries (approximately 3,300 records), focusing on privacy protection and content fairness.

\subsubsection{Privacy Protection Verification}
The privacy protection verification process systematically examines potential privacy concerns within the dataset. Table~\ref{tab:privacy_criteria} outlines our evaluation criteria and standards.

\begin{table}[t]
\centering
\small
\begin{tabular}{p{0.18\linewidth}|p{0.36\linewidth}|p{0.27\linewidth}}
\toprule
\textbf{Aspect} & \textbf{Verification Content} & \textbf{Acceptance Criteria} \\
\midrule
Personal Identity & Names, addresses, contact information & Strictly prohibited \\
\midrule
Indirect Identifiers & Combinations of information that could lead to identification & Must not enable personal identification \\
\midrule
Sensitive Data & Health conditions, financial details & Limited to general product-related information \\
\bottomrule
\end{tabular}
\caption{Privacy protection verification criteria for the PDD dataset}
\label{tab:privacy_criteria}
\end{table}

\subsubsection{Fairness Assessment}
Our fairness assessment framework examines potential biases and discriminatory content within the dataset. This evaluation ensures that product descriptions maintain objectivity and avoid perpetuating societal stereotypes. Table~\ref{tab:fairness_criteria} presents our fairness evaluation framework.

\begin{table}[t]
\centering
\small
\begin{tabular}{p{0.18\linewidth}|p{0.36\linewidth}|p{0.27\linewidth}}
\toprule
\textbf{Category} & \textbf{Assessment Focus} & \textbf{Requirements} \\
\midrule
Gender & Gender-related stereotypes and biases & Neutral product descriptions without gender discrimination \\
\midrule
Ethnicity & Racial or ethnic biases & No ethnicity-specific stereotypes or prejudices \\
\midrule
Cultural Elements & Cultural sensitivity and representation & Objective and culturally neutral descriptions \\
\bottomrule
\end{tabular}
\caption{Fairness assessment criteria for dataset evaluation}
\label{tab:fairness_criteria}
\end{table}

\subsection{Model Output Evaluation}
The evaluation of model-generated product descriptions focuses on two fundamental constraints: completeness and accuracy. We randomly selected 1,000 samples from the test set for this assessment, with three domain experts conducting independent evaluations.

\subsubsection{Evaluation Methodology}
Our evaluation methodology employs a binary scoring system (0 or 1) based on strict compliance with both completeness and accuracy requirements. Table~\ref{tab:eval_criteria} details our scoring criteria.

\begin{table}[t]
\centering
\small
\begin{tabular}{p{0.135\linewidth}|p{0.315\linewidth}|p{0.36\linewidth}}
\toprule
\textbf{Score} & \textbf{Requirements} & \textbf{Assessment Criteria} \\
\midrule
1 & Complete satisfaction of all constraints & All property-value pairs included; No additional information introduced \\
\midrule
0 & Failure to meet any constraint & Missing any property-value pair OR Including extraneous information \\
\bottomrule
\end{tabular}
\caption{Model output evaluation criteria and scoring system}
\label{tab:eval_criteria}
\end{table}

\subsubsection{Evaluation Protocol}
The evaluation protocol ensures consistency and reliability across assessments. Each evaluator independently examines the generated descriptions, comparing them against the input property-value pairs. For quality control, we conducted preliminary training sessions and established a standardized evaluation process. Disagreements among evaluators were resolved through detailed discussion and consensus building.

The final evaluation score for each generated description represents the average of scores from all evaluators. To ensure evaluation reliability, we calculated the inter-rater agreement using Cohen's Kappa coefficient. For cases receiving a score of 0, evaluators documented specific violation types, enabling detailed analysis of model limitations and potential areas for improvement.

\section{Training Expense}
\label{app:training}
\subsection{Computing Infrastructure}
All experiments in this study were conducted using the computing resources detailed in Table~\ref{tab:compute-resources}. To ensure reproducibility and consistent performance, we utilized the same hardware for all evaluations and training.

\begin{table}[t]
\centering
\small
\begin{tabular}{ll}
\toprule
\textbf{Component} & \textbf{Specification} \\
\midrule
CPU & Intel Xeon E5-2680 v4 @ 2.40GHz \\
RAM & 128GB DDR4 \\
GPU & NVIDIA A100 80GB \\
Operating System & Ubuntu 20.04 LTS \\
CUDA Version & 12.1 \\
Python Version & 3.9.12 \\
\bottomrule
\end{tabular}
\caption{Computing Infrastructure Specifications}
\label{tab:compute-resources}
\end{table}

\subsection{Training Configuration}
All hyperparameter settings are listed in Table~\ref{tab:training_params}.
Given that IFEval contains only 430 training samples, we adopted smaller batch sizes and larger initial learning rates when training on the IFEval dataset.

\subsection{Generation Configuration}
For our experiments, we employed carefully selected parameters to ensure consistent and reproducible results, as shown in Table~\ref{tab:model-params}. These parameters were chosen to minimize output variability while maintaining generation quality.

\begin{table}[t]
\centering
\small
\begin{tabular}{ll}
\toprule
\textbf{Parameter} & \textbf{Value} \\
\midrule
Temperature & 0.0 \\
Top P & 1.0 \\
Frequency Penalty & 0.0 \\
Presence Penalty & 0.0 \\
Maximum Tokens & 2048 \\
Context Window & 16385 tokens \\
\bottomrule
\end{tabular}
\caption{Language Model Parameters}
\label{tab:model-params}
\end{table}

The temperature was set to 0.0 to maximize deterministic behavior, while maintaining a top P value of 1.0 to preserve the model's ability to generate coherent responses. Both frequency and presence penalties were set to 0.0 to avoid artificial constraints on the model's token selection process. These settings were kept constant across all experiments to ensure consistent generation behavior and reproducible results.

\begin{table}[h]
\centering
\resizebox{\columnwidth}{!}{
\begin{tabular}{|l|c|c|}
\hline
\textbf{Parameter} & \textbf{PDD Dataset} & \textbf{IFEval Dataset} \\
\hline
Training Samples & 3300 & 430 \\
\hline
\multicolumn{3}{|l|}{\textbf{SFT}} \\
\hline
Learning Rate & 5e-6 & 1e-4 \\
Train Batch Size & 256 & 32 \\
Micro Train Batch Size & 4 & 4 \\
Max Sequence Length & 4096 & 4096 \\
Max Epochs & 2 & 2 \\
\hline
\multicolumn{3}{|l|}{\textbf{DPO}} \\
\hline
Learning Rate & 5e-7 & 1e-4 \\
Train Batch Size & 128 & 32 \\
Micro Train Batch Size & 4 & 4 \\
Max Sequence Length & 4096 & 4096 \\
Max Epochs & 2 & 2 \\
Beta & 0.1 & 0.1 \\
\hline
\multicolumn{3}{|l|}{\textbf{KTO}} \\
\hline
Learning Rate & 5e-7 & 1e-4 \\
Train Batch Size & 128 & 32 \\
Micro Train Batch Size & 4 & 4 \\
Max Sequence Length & 4096 & 4096 \\
Max Epochs & 2 & 2 \\
Beta & 0.1 & 0.1 \\
\hline
\multicolumn{3}{|l|}{\textbf{SimPO}} \\
\hline
Learning Rate & 5e-7 & 1e-4 \\
Train Batch Size & 128 & 32 \\
Micro Train Batch Size & 4 & 4 \\
Max Sequence Length & 4096 & 4096 \\
Max Epochs & 2 & 2 \\
Beta & 0.1 & 0.1 \\
\hline
\multicolumn{3}{|l|}{\textbf{ORPO}} \\
\hline
Learning Rate & 5e-7 & 1e-4 \\
Train Batch Size & 128 & 32 \\
Micro Train Batch Size & 4 & 4 \\
Max Sequence Length & 4096 & 4096 \\
Max Epochs & 2 & 2 \\
Beta & 0.1 & 0.1 \\
\hline
\multicolumn{3}{|l|}{\textbf{PPO}} \\
\hline
Actor Learning Rate & 5e-7 & 1e-4 \\
Critic Learning Rate & 9e-6 & 2e-4 \\
Train Batch Size & 128 & 32 \\
Micro Train Batch Size & 2 & 2 \\
Rollout Batch Size & 1024 & 1024 \\
Micro Rollout Batch Size & 4 & 4 \\
Max Epochs & 2 & 2 \\
KL Coefficient & 0.01 & 0.01 \\
Max Prompt Length & 1024 & 1024 \\
Max Generate Length & 3072 & 3072 \\
\hline
\multicolumn{3}{|l|}{\textbf{GAPO}} \\
\hline
Actor Learning Rate & 5e-7 & 1e-4 \\
Critic Learning Rate & 9e-6 & 2e-4 \\
Train Batch Size & 128 & 16 \\
Micro Train Batch Size & 2 & 2 \\
Rollout Batch Size & 1024 & 1024 \\
Micro Rollout Batch Size & 4 & 4 \\
Classifier Batch Size & 8 & 4 \\
Classifier Learning Rate & 1e-5 & 1e-5 \\
Max Prompt Length & 1024 & 1024 \\
Max Generate Length & 3072 & 3072 \\
KL Coefficient & 0.01 & 0.01 \\
Adversarial Training Epochs & 2 & 2 \\
Classifier Warmup Epochs & 2 & 2 \\
Classifier Training Epochs & 2 & 2 \\
Max Epochs & 2 & 2 \\
Classifier Generator Ratio & 0.5 & 0.5 \\
\hline
\end{tabular}
}
\caption{Hyperparameter Settings for Different Training Methods}
\label{tab:training_params}
\end{table}

\section{Prompt}
\label{app:prompt}
We use a comprehensive prompt template as shown in Table~\ref{tab:product-description-prompt}.
The template includes essential components such as product name, word count requirement, emotion specifications, and factual information.
To explore the performance of different prompt engineers strategies, we further implement three distinct output formats (Table~\ref{tab:prompt}), namely Naive, Chain-of-Thought (CoT), and Plan-N-Solve approaches.

\begin{table*}[t]
    \centering
    \begin{tabular}{|p{0.95\textwidth}|}
    \hline
\# -*- coding: utf-8 -*-\\
Variables:\\
!<INPUT 0>! -- Product Name\\
!<INPUT 1>! -- Word Count Requirement\\
!<INPUT 2>! -- Emotion Type and Description\\
!<INPUT 3>! -- Factual Information\\
!<INPUT 4>! -- Output Instruction\\
<commentblockmarker>\#\#\#</commentblockmarker>\\
Please generate a product description about !<INPUT 0>! with approximately !<INPUT 1>! words.\\
You need to use all the information provided in the Factual Information section to generate the description. The description should convey the emotion specified in the Emotion section. \\
Note that you cannot add additional factual information, and you must use all the given facts. Please only add non-factual, emotion-related content.\\\\
\#\#\# Emotion:\\
!<INPUT 2>!\\\\
\#\#\# Factual Information:\\
!<INPUT 3>!\\\\
\#\#\# Your output should follow this format:\\
!<INPUT 4>!\\
\hline
\end{tabular}

\caption{Base template for the experiment of product description generation in this paper.}
\label{tab:product-description-prompt}
\end{table*}

\begin{table*}[t]
\centering
\resizebox{\textwidth}{!}{
\renewcommand{\arraystretch}{1.3}
\begin{tabular}{|l|p{0.8\textwidth}|}
\hline
\rowcolor{gray!10} \textbf{Method} & \textbf{Prompt Template} \\
\hline
\texttt{Naive} & !<INPUT 4>! = \\
& The description should be generated below the ``\#\#\# Generated Result:'' \\
\hline
\texttt{CoT} & !<INPUT 4>! = \\
& Generate your thinking process step by step below the ``\#\#\# Thinking Process:'' \\
& Then the description should be generated below the ``\#\#\# Generated Result:'' \\
\hline
\texttt{Plan-N-Solve} & !<INPUT 4>! = \\
& Generate your planing step by step below the ``\#\#\# Planning:'' \\
& Then the description should be generated below the ``\#\#\# Generated Result:'' \\
\hline
\end{tabular}
}
\caption{Detail prompt request in Tab.~\ref{tab:product-description-prompt}.}
\label{tab:prompt}
\end{table*}

\end{CJK}
\end{document}